\documentclass[12pt]{article}
\usepackage{amsmath}
\usepackage{graphicx}
\usepackage{geometry}

\usepackage{natbib}

\ifx\pdfoutput\@undefined\usepackage[usenames,dvips]{color}
\else\usepackage[usenames,dvipsnames]{color}
\IfFileExists{pdfcolmk.sty}{\usepackage{pdfcolmk}}{} 
\fi
\usepackage[plainpages=false,pdfpagelabels,pagebackref=false,naturalnames=true,hyperindex=true,pdftitle={What Does Artificial Life Tell Us About Death?},pdfauthor={Carlos Gershenson}]{hyperref}
\hypersetup{colorlinks=true,
urlcolor=Cerulean,linkcolor=BrickRed,citecolor=RoyalBlue,a4paper,
  pdfpagemode=None,
  pdfstartview=FitH}

\begin{document}

\title{What Does Artificial Life Tell Us About Death?}
\author{Carlos Gershenson$^{1,2,3}$ \\
$^{1}$ Instituto de Investigaciones en Matem\'aticas Aplicadas y en Sistemas \\
Universidad Nacional Aut\'onoma de M\'exico\\
Ciudad Universitaria\\
Apdo. Postal 20-726 / Adm—n No. 20\\
01000 M\'exico D.F. M\'exico\\
\href{mailto:cgg@unam.mx}{cgg@unam.mx} \
\url{http://turing.iimas.unam.mx/~cgg} \\
$^{2}$ Centro de Ciencias de la Complejidad \\
Universidad Nacional Aut\'onoma de M\'exico\\
$^{3}$Centrum Leo Apostel, Vrije Universiteit Brussel\\
Krijgskundestraat 33 B-1160 Brussel, Belgium\\
\href{mailto:cgershen@vub.ac.be}{cgershen@vub.ac.be} \ \url{http://homepages.vub.ac.be/~cgershen}}
\maketitle

%


\begin{quote}
\begin{flushright}
\textit{Every evil leaves a sorrow in the memory,\\
 until the supreme evil, death,\\
 wipes out all memories together with all life.}

\textit{--Leonardo da Vinci }
\end{flushright} 
\end{quote}

One of the open problems in artificial life discussed by Bedau, et al. \citep{BedauEtAl2000} is the establishment of ethical principles for artificial life. In particular: 
\begin{quote}
Much of current ethics is based on the sanctity of human life. Research in artiÞcial life will affect our understanding of life and death (...) This, like the theory of evolution, will have major social consequences 
for human cultural practices such as religion.  \cite[p. 375]{BedauEtAl2000}
\end{quote}

Focussing on our understanding of death, this will depend necessarily on our understanding of life, and vice versa. Throughout history there have been several explanations to both life and death, and it seems unfeasible that a consensus will be reached. Thus, we are faced with multiple notions of life, which imply different notions of death. However, generally speaking, if we describe life as a process, death can be understood as the irreversible termination of that process. 

The general notion of life as a process or organization \citep{Langton:1989,Sterelny:1999,Korzeniewski:2001} has expelled vitalism from scientific worldviews. Moreover, there are advantages in describing living systems from a functional perspective, e.g. it makes the notion of life independent of its implementation. This is crucial for artificial life. Also, we know that there is a constant flow of matter and energy in living systems, i.e. their physical components can change while the identity of the organism is preserved. In this respect, one can make a variation of  Kauffman's ``blender thought experiment" \citep{Kauffman2000}: if you put a macroscopic living system in a blender and press ``on", after some seconds you will have the same molecules that the living system had. However, the organization of the living system is destroyed in the blending. Thus, life is an organizational aspect of living systems, not so much a physical aspect. Death occurs when this organization is lost.

One of the main properties of living organization is its self-production \citep{VarelaEtAl1974,MaturanaVarela1980,MaturanaVarela1987,Luisi:2004,Kauffman2000}. When death occurs, this self-production cannot be maintained. But is this organization the only thing that is lost with death?

The notions of life and death have been much related to those of mind, cognition, awareness, consciousness, and soul\footnote{It is quite problematic to attempt to define these, but a vague notion will suffice. In the following, ``mind" will be used in a broad sense that includes also cognition, awareness, consciousness, and soul.}. On the one hand, the mind is a property closely related with life. Some propose that mind and life are essentially the same process \citep{Stewart:1996,Bedau1998}. On the other hand, people have speculated since the dawn of civilization on what occurs with the mind after death.

Life is a process described by an observer \citep{MaturanaVarela1987}, in first or third person perspective. When the process breaks, only description in the third person observer remains. By definition, we can only speak about death from a third person perspective.

What can artificial life add to this discussion? Artificial life simulations (``soft" ALife) can be seen as opaque thought experiments \citep{Di-Paolo:2000}, i.e. one can explore different notions of life and death with them. Robots (``hard" ALife) would also serve this purpose. Artificial life can help us build living systems to be explored from a third person perspective in a synthetic way \citep{Steels1993}.
Can we say that ``animats" \citep{Wilson:1985} have a mind, in the same sense as animals do? If not, is there something missing in the particular animat, in artificial life, or in the observer? When a digital organism dies \citep{Ray1994}, what physically changes is the RAM that encoded the organism in bits. When the bits describing the organization of the organism are erased, the only place where the organism prevails is in the observer. The same is for robots. The same is for animals. The same is for humans. If we describe life as an organizational process, and a mind as depending on it \citep{Clark1997}, when the organization is lost, the life is lost and the mind is lost.

Certainly, the organization of digital organisms is much easier to preserve than that of biological ones. Apart from the ease of copying digital information, digital organisms are generally inhabiting closed environments. Biological organisms face open dynamical environments that constantly threaten their integrity. i.e. organisms need to make thermodynamical work (metabolism) \citep{Kauffman2000} to maintain themselves. In an open environment such as the biosphere, where different evolving organisms interact, there is no ``best" or ``fittest" organism, since the fitness depends on the dynamic environment. Thus, fitness changes constantly with the environment, since the environment is changed as organisms evolve trying to increase their fitness. In this context, it can be speculated that there is an evolutionary advantage of death. If there was no death, i.e. if an organism somehow managed to maintain its organization indefinitely, evolution would stop. This actually occurs commonly with digital evolution. In fact, death of digital organisms has been used as a measure to introduce novelty \citep{Ray1994,Dorin:2005,OlsenEtAl2008}.

The organization represented in a species can survive several lifespans, but is subject to the same pressure as the one just described. The loss of organization gives the opportunity to new forms of organization to develop.

The development of protocells \citep{Protocells2008} (``wet" ALife) might further contribute to the exploration of the notions of life and death. By chemically producing the organization resulting in living systems, the non-mystical notion of life reviewed here will gain further grounds. Additionally, the non-mystical notion of death explored here will have to be further elaborated. What occurs when a protocell dies? If we can create again a living system with the same organization, did it die in the first place? We will be able to have different instantiations of the same living organization, just like we can have different copies of the same digital organism. Will its death have the same meaning as that of an animal?

One thing to notice in these questions is that in most biological organisms, the organization lies not only in their genes, but also in their development (epigenesis). Clones can develop different organizations. The same might occur for protocells and other future ``wet" artificial living systems. However, on the digital side of artificial life, it is easy again to maintain and reproduce the organization acquired through development \citep{EpiRob2001}.

What will the future bring? Will there be biological systems closer to digital ones, in the sense that living information can be maintained and/or reproduced? Probably. How will this affect death? We will have more control over it. Will this mark an end to evolution? No, even when some living organization might be more persistant, there will always be new situations where organisms have to adapt. In any case, the cultural attitudes towards death most probably will change. This is not suggesting that we will be less touched by it, or less spiritual towards it. The implication is that we will have a better understanding of the phenomenon, with a broader scientific basis.

To conclude this philosphical essay, different notions of death will be deduced from a limited set of different notions of life:

\begin{itemize}
\item If we consider life as self-production \citep{VarelaEtAl1974,MaturanaVarela1980,MaturanaVarela1987,Luisi:2004}, then death will the the loss of that self-production ability.
\item If we consider life as what is common to all living beings \cite[p. 8]{DeDuve2003}, then death implies the termination of that commonality, distinguishing it from other living beings. 
\item If we consider life as computation \citep{Hopfield1994}, then death will be the end (halting?) of that computing process.
\item  If we consider life as supple adaptation \citep{Bedau1998}, death implies the loss of that adaptation.
\item  If we consider life as a self-reproducing system capable of at least one thermodynamic work cycle  \cite[p. 4]{Kauffman2000}, death will occur when the system will be unable to perform thermodynamic work. \item If we consider life as information (a system) that produces more of its own information than that produced by its environment \citep{Gershenson:2007}, then death will occur when the environment will produce more information than that produced by the system.
\end{itemize}





\bibliographystyle{cgg}
\bibliography{carlos,evolution,RBN,information,COG}

\end{document}